\documentclass[runningheads]{llncs}
\usepackage[T1]{fontenc}
\usepackage{graphicx,verbatim}
\usepackage{xcolor}
\usepackage{xcolor}
\usepackage[T1]{fontenc}
\usepackage{microtype}
\usepackage{amsmath}
\usepackage{amssymb}
\begin{document}
%
\title{TTDF: A Two-Stage Framework for Reliable Surgical Phase Transition Detection}
\titlerunning{TTDF for Reliable Surgical Phase Transition Detection}
%

\author{Yushi Guo\inst{1,3}\thanks{Corresponding author}\and
Pietro Valdastri\inst{2,3}\and
Duygu Sarikaya\inst{1,3}}  
\authorrunning{Y. Guo et al.}
\institute{School of Computer Science, University of Leeds, Leeds, UK
\and
School of Electronic and Electrical Engineering, University of Leeds, Leeds, UK
\and
STORM Lab UK \\
    \email{scyg@leeds.ac.uk}}
  
\maketitle              
\begin{abstract}
Reliable workflow transition detection is important for context-aware surgical assistance and downstream decision support. However, online surgical phase recognizers primarily focus on frame-wise accuracy and temporal consistency, rather than the reliability of workflow transition events. Directly converting phase changes into events is unreliable: temporal jitter and workflow-illegal switches produce false or duplicate events, while persistent, workflow-consistent candidates may remain incorrect. To address this limitation, we formulate reliable workflow transition detection as a distinct event-level task operating on outputs of a frozen online phase recognizer. We propose the Two-Stage Transition Detection Framework (TTDF), a causal framework that progressively filters transition candidates. Transition Candidate Extraction (TCE) first applies a minimum-duration requirement and a workflow-graph constraint to remove false candidates caused by temporal jitter and phase transitions not allowed by the workflow graph. Specifically, a candidate is retained only if the predicted target phase persists for a minimum duration and the ordered phase pair belongs to the workflow graph’s allowed transition set. TCE thereby produces a high-recall candidate set without additional training. Transition Candidate Verification (TCV) suppresses remaining false candidates using phase-posterior shifts and visual-change cues from frozen DINOv2 features. Events are assessed using a phase-pair-aware one-to-one matching protocol. Experiments on Cholec80 show that TTDF reduces false transition emissions while preserving recall and controlling decision delay.
\end{abstract}


%
%
%
\section{Introduction}
Surgical phase recognition plays a critical role in enhancing intraoperative situation awareness, enabling real-time decision support and post-operative analysis, and supporting context-aware computer-assisted intervention (CAI) applications such as monitoring the surgical process~\cite{franke2018intelligent}, operating-room
coordination, and phase-dependent decision support for surgical training~\cite{wagner2023comparative}. 
Existing surgical phase recognition methods primarily improve dense
frame-wise prediction by strengthening temporal context modeling.
To this end, hierarchical transformer architectures, such as MuST~\cite{perez2024must}
and Surgformer~\cite{yang2024surgformer}, capture long-range workflow dependencies and improve temporal consistency across surgical phases. Complementing these approaches, boundary-aware methods explicitly model phase-transition regions. For example, LoViT~\cite{liu2025lovit} introduces transition-region
supervision to reduce prediction jitter around phase boundaries.
More recently, several studies have investigated the temporal reliability of online surgical phase recognition. Liu et al.~\cite{liu2026stabilizing} reduce prediction
fragmentation through temporal error-cascade training and evidence-gated phase switching, while DSTED~\cite{chen2026dsted} combines reliable memory propagation with uncertainty-aware prototype retrieval to stabilize online predictions and improve ambiguous-phase discrimination. Despite these advances, existing methods still treat phase changes mainly as consequences of dense phase recognition and evaluate performance at the frame or segment level.

Directly converting online phase-label changes into workflow transition events
is unreliable. Temporal jitter can generate false or duplicate events, while some
predicted transitions are not permitted by the workflow graph. Minimum-duration
filtering and workflow legality remove many such candidates. However, some remain
temporally persistent and workflow-consistent yet still incorrect, and cannot be
reliably resolved by these constraints alone. This motivates Transition Candidate
Verification using phase-posterior and visual evidence. Consequently, high frame-wise
phase accuracy and temporal stability do not guarantee reliable transition event
emission for downstream decision support.


To address the distinct failure modes of raw phase changes, we propose the Two-Stage Transition Detection Framework (TTDF), a causal framework operating on the outputs of a frozen online phase recognizer. Transition Candidate
Extraction (TCE) first removes candidates caused by temporal jitter and disallowed
phase transitions using dwell persistence and workflow legality, producing a
high-recall set of persistent, workflow-consistent candidates without additional
training. Transition Candidate Verification (TCV) subsequently suppresses remaining
false candidates using local phase-posterior shifts and visual-change cues within a
causal candidate-centered window. Accordingly, TCE removes temporal jitter and workflow-illegal transitions through training-free candidate filtering, whereas TCV removes persistent, workflow-consistent false candidates through learned candidate filtering. \\
We evaluate the resulting workflow transition events using an event-level evaluation protocol based on phase-pair-aware one-to-one temporal matching. Unlike frame-wise and segment-level evaluation, the proposed protocol assesses the correctness of the ordered phase pair while explicitly accounting for false events, duplicate events, transition-time localization, and online decision delay. Experiments on Cholec80~\cite{twinanda2016endonet} demonstrate that the proposed two-stage error-reduction strategy reduces false transition emissions while preserving event recall and maintaining controlled online decision delay. Our main contributions are summarized as follows:
\begin{itemize}
    \item We formulate reliable workflow transition detection as an event-level
    task distinct from online surgical phase recognition, with each prediction
    represented by an ordered phase pair and a transition time.
    
    \item We propose TTDF, a lightweight two-stage framework that progressively removes false candidates caused by temporal jitter and phase transitions not allowed by the workflow graph, followed by verification of the remaining false candidates from the outputs of a frozen online phase recognizer.
    
    \item We adopt a phase-pair-aware event-level evaluation protocol that measures transition detection accuracy, false and duplicate detections, temporal localization error, and online decision delay.
\end{itemize}

\section{Methods}

\subsection{Problem Formulation and Framework Overview}

Given an online surgical video
$\mathcal{V}=\{v_t\}_{t=1}^{T}$, a frozen online phase recognizer produces
at each time step a phase-posterior vector
$\mathbf{p}_t\in[0,1]^C$ and the corresponding hard phase prediction
$\hat{y}_t=\arg\max_c\mathbf{p}_t(c)$. Rather than modifying the
frame-wise phase predictions, our objective is to detect discrete
workflow transition events from these causal recognizer outputs. Each event
indicates that the procedure has transitioned from a source phase $A$ to a
target phase $B$.

A transition candidate is represented as
\begin{equation}
c_i=
(A_i,B_i,t_i^{\mathrm{tr}},t_i^{\mathrm{dec}}),
\end{equation}
where $t_i^{\mathrm{tr}}$ denotes the first predicted transition from phase
$A_i$ to phase $B_i$, and $t_i^{\mathrm{dec}}$ denotes the time at which the final retain-suppress decision is produced by the complete two-stage
framework. Stage 1 first confirms that the candidate satisfies the dwell and workflow constraints, after which Stage 2 evaluates the retained candidate using candidate-centered posterior and visual evidence.

For each candidate, the framework produces a binary decision
$d_i\in\{0,1\}$, where $d_i=1$ denotes retain and $d_i=0$ denotes suppress. The final emitted event set is
\begin{equation}
\hat{\mathcal{E}}
=
\left\{
(A_i,B_i,t_i^{\mathrm{tr}},t_i^{\mathrm{dec}})
\mid d_i=1
\right\}.
\end{equation}
Thus, $t_i^{\mathrm{tr}}$ specifies the estimated transition location, whereas
$t_i^{\mathrm{dec}}$ records when the final decision becomes available to an
online downstream system. Their difference,
$t_i^{\mathrm{dec}}-t_i^{\mathrm{tr}}$,defines the candidate confirmation latency.

We characterize transition reliability through ordered phase-pair correctness, temporal correspondence to the annotated event, low false and duplicate
emissions, and bounded online decision delay. These event-level properties are not fully captured by conventional frame-wise or segment-level phase recognition metrics.
Raw changes in the hard phase predictions exhibit two complementary error regimes. First, temporal jitter can generate false or duplicate transition candidates, while workflow-illegal phase transitions can produce invalid phase-pair transitions. Errors arising from these two sources can be removed or substantially reduced using dwell persistence and workflow legality. However, some candidates remain temporally persistent and workflow-legal but still do not correspond to true workflow transitions, owing to an incorrect ordered phase pair or inaccurate transition timing.
These remaining false candidates therefore require candidate-centered classification using local phase-posterior shifts and visual-change cues.

\begin{figure*}[t]
    \centering
    \includegraphics[
        width=\textwidth,
        trim=25 10 25 10,
        clip
    ]{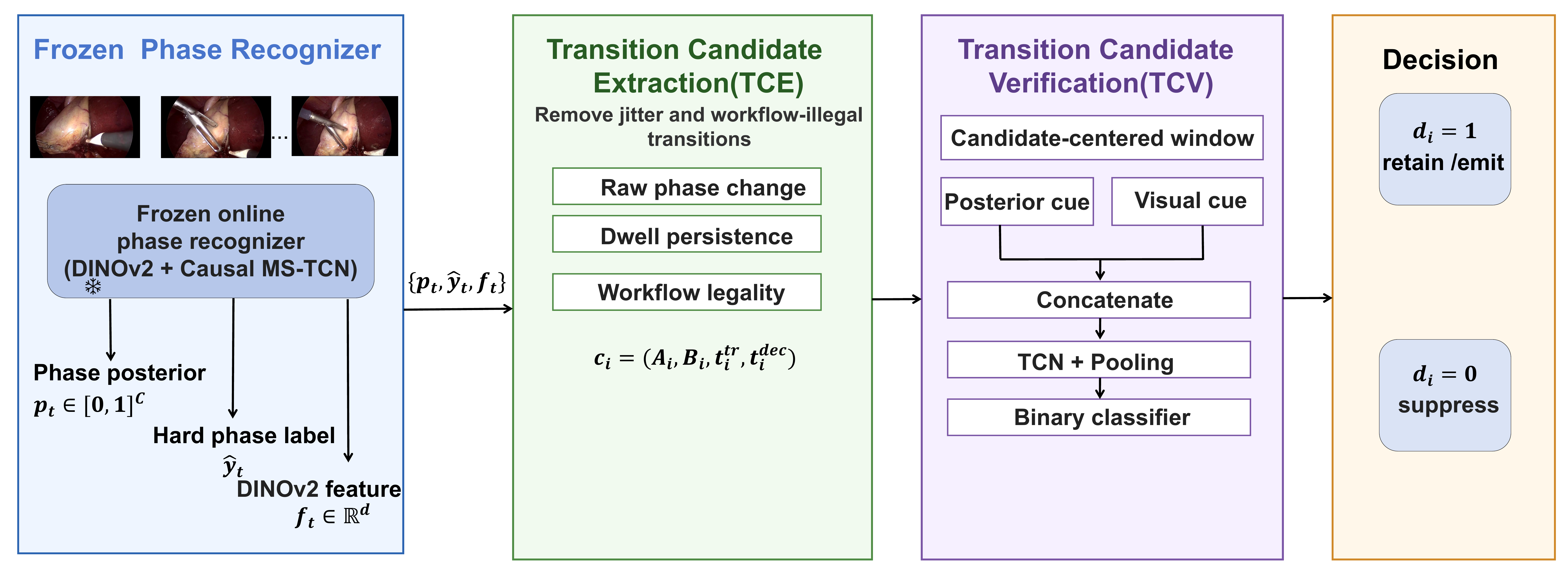}
    
    \caption{
    Overview of the proposed Two-Stage Transition Detection Framework (TTDF). Given the online phase predictions, the Transition Candidate Extraction (TCE) module first removes false candidates caused by temporal jitter and phase transitions not allowed by the workflow graph, thereby constructing a high-recall transition candidate set. The Transition Candidate Verification (TCV) module then verifies each candidate using candidate-centered temporal representations and local visual-change cues, producing the final transition decisions.
    }
    \label{fig:framework}
\end{figure*}

We propose the \emph{Two-Stage Transition Detection Framework} (TTDF), a causal framework operating on a frozen online phase recognizer. \emph{Transition Candidate Extraction} (TCE) first removes false candidates caused by temporal jitter and disallowed phase transitions, constructing a high-recall set of persistent, workflow-consistent candidates without additional training.
\emph{Transition Candidate Verification} (TCV) then suppresses remaining false candidates using local phase-posterior shifts and visual-change cues. Formally, the final decision is
\begin{equation}
d_i = \mathbf{1}\left[\sigma\left(g_\theta(\phi_i)\right) \geq \gamma\right],
\end{equation}
where $\phi_i$ denotes the encoded candidate-centered cue representation,
$g_\theta$ denotes the binary scoring head that produces a scalar logit,
$\sigma(\cdot)$ is the sigmoid function, and $\gamma$ is selected on the
validation set and fixed during testing. An overview of TTDF is shown in Fig.~\ref{fig:framework}.

\subsection{Transition Candidate Extraction}
Transition Candidate Extraction (TCE) is a training-free module that constructs
a high-recall transition candidate set from the frozen hard phase predictions
described in Sec.~2.1. It removes false candidates caused by temporal jitter
and phase transitions not allowed by the workflow graph. In our implementation,
the upstream online phase recognizer
consists of frozen DINOv2~\cite{oquab2023dinov2} features followed by a causal MS-TCN adapted from TeCNO~\cite{czempiel2020tecno}. During online inference, the $i$-th raw transition candidate is generated
whenever the prediction changes from phase $A_i$ to phase $B_i$:
\[
\hat y_{t_i^{tr}-1}=A_i,\quad
\hat y_{t_i^{tr}}=B_i,\quad
A_i\neq B_i.
\]

This condition defines a raw transition candidate rather than a final
workflow-transition event, since short-lived prediction fluctuations may
produce spurious phase changes. 
Let $t_i^{tr}$ denote the first time at which the prediction changes from
$A_i$ to $B_i$.

To suppress temporal jitter, the candidate is retained only if the target
phase remains stable for a dwell duration $d$,

\begin{equation}
\hat{y}_t = B_i,\quad t=t_i^{\mathrm{tr}},\ldots,t_i^{\mathrm{tr}}+d-1 .
\end{equation}
To further remove structurally invalid transitions, the ordered phase pair is required to belong to the workflow graph estimated exclusively from the training set,
\[
(A_i,B_i)\in \mathcal{E}_{train}.
\] 
Each retained candidate is represented as

\[
c_i=
(A_i,B_i,
t_i^{\mathrm{tr}},
t_i^{\mathrm{dec}}),
\]

where $t_i^{\mathrm{dec}}$ denotes the time at which the final retain-suppress decision is produced by the complete two-stage framework.

After applying minimum-duration filtering and the workflow-graph constraint,
most false candidates caused by temporal jitter and phase transitions not
allowed by the workflow graph are removed while preserving high recall.
However, some candidates remain temporally persistent and workflow-consistent
but still correspond to incorrect workflow transitions. These remaining false
candidates are handled by the proposed Transition Candidate Verification (TCV).

\subsection{Transition Candidate Verification}

\begin{figure}[!t]
    \centering
    \includegraphics[
        width=0.82\textwidth,
        trim=25 5 25 10,
        clip
    ]{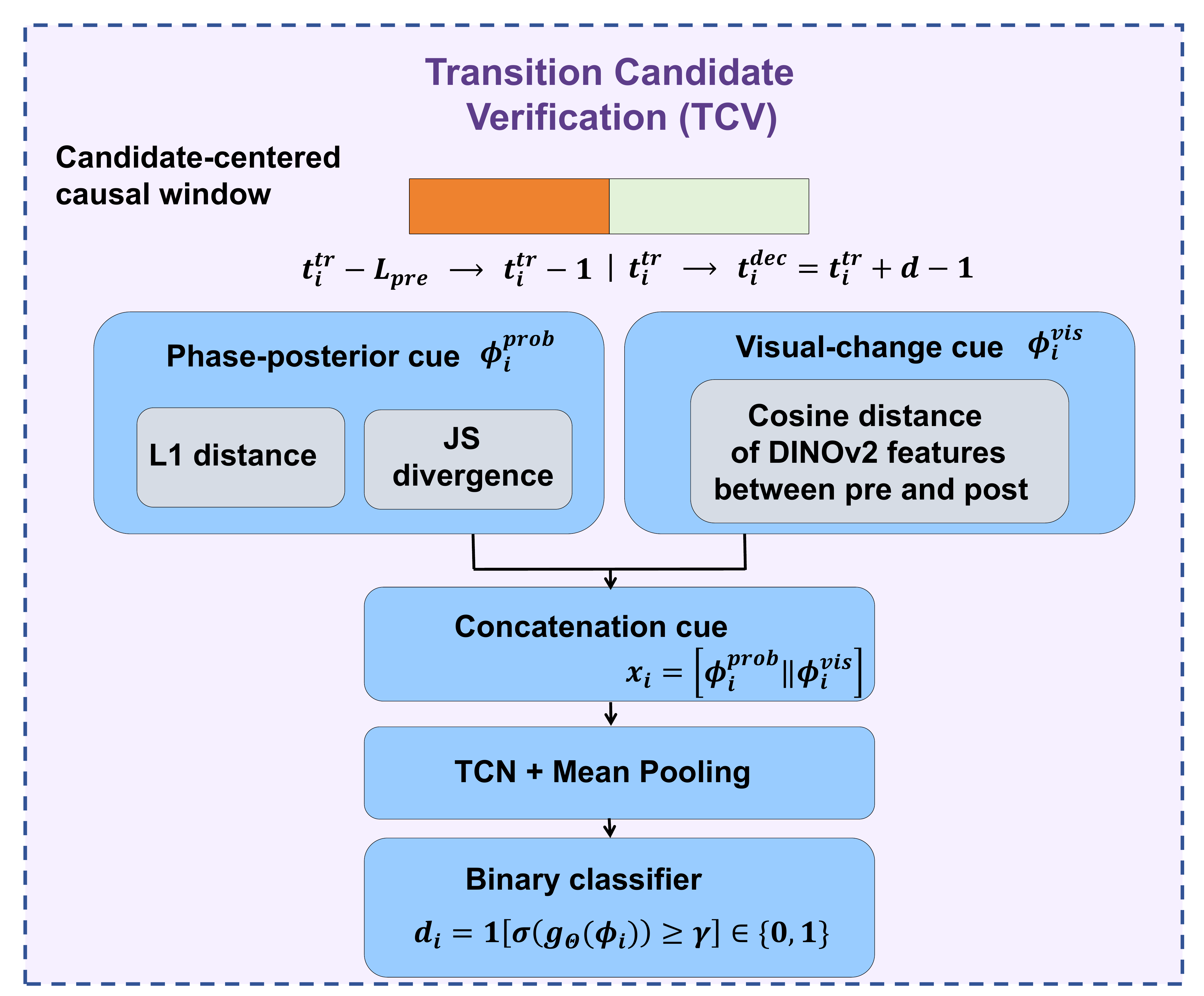}
    \caption{Detailed illustration of the Transition Candidate Verification (TCV) module. Posterior and visual-change cues from a causal candidate-centered window are concatenated and processed by a lightweight TCN with mean pooling for binary verification.}
    \label{fig:tcv}
\end{figure}

Although Transition Candidate Extraction removes many false candidates caused by temporal jitter and phase transitions not allowed by the workflow graph, some retained candidates
remain false after passing dwell-persistence and workflow-legality filtering.
These remaining false candidates require learned candidate filtering based on the
local posterior and visual evidence associated with each transition candidate.
Specifically, a valid transition candidate is expected to exhibit a coherent
change in both the phase-posterior distribution and the visual representation
across the candidate boundary. We introduce Transition Candidate Verification (TCV), which integrates these complementary cues and
predicts a binary retention decision for each candidate generated by TCE. The overall TCV process is illustrated in Fig.~\ref{fig:tcv}.

For each candidate $c_i$, we construct a causal candidate-centered window
anchored at its transition time $t_i^{\mathrm{tr}}$. The pre-transition interval
contains observations immediately preceding $t_i^{\mathrm{tr}}$, whereas the
post-transition interval spans from $t_i^{\mathrm{tr}}$ to the end of the dwell
interval. Once this causal window becomes available, TCV produces the binary
decision $d_i$ at decision time $t_i^{\mathrm{dec}}$.

\paragraph{Phase-posterior distribution cue.}
Let $\bar{\mathbf{p}}_i^{\mathrm{pre}}$ and
$\bar{\mathbf{p}}_i^{\mathrm{post}}$ denote the mean phase-posterior vectors
within the pre- and post-transition windows, respectively. We characterize
the posterior change as
\begin{equation}
\boldsymbol{\phi}_i^{\mathrm{prob}}
=
\left[
\left\|
\bar{\mathbf{p}}_i^{\mathrm{post}}
-
\bar{\mathbf{p}}_i^{\mathrm{pre}}
\right\|_1,
\;
D_{\mathrm{JS}}
\left(
\bar{\mathbf{p}}_i^{\mathrm{pre}}
\parallel
\bar{\mathbf{p}}_i^{\mathrm{post}}
\right)
\right],
\end{equation}
where the $\ell_1$ distance measures the magnitude of the posterior change,
and $D_{\mathrm{JS}}$ denotes the Jensen--Shannon divergence~\cite{lin1991divergence} 
between the phase-posterior distributions on the two sides of the candidate
transition.

\paragraph{Visual-change cue.}
To provide complementary visual evidence, we use the frozen DINOv2 \cite{oquab2023dinov2} features
from the same temporal windows. Let
$\bar{\mathbf{f}}_i^{\mathrm{pre}}$ and
$\bar{\mathbf{f}}_i^{\mathrm{post}}$ denote their mean feature vectors.
The visual-change cue is defined as
\begin{equation}
\phi_i^{\mathrm{vis}}
=
1-
\frac{
\left\langle
\bar{\mathbf{f}}_i^{\mathrm{pre}},
\bar{\mathbf{f}}_i^{\mathrm{post}}
\right\rangle
}{
\left\|
\bar{\mathbf{f}}_i^{\mathrm{pre}}
\right\|_2
\left\|
\bar{\mathbf{f}}_i^{\mathrm{post}}
\right\|_2
}.
\end{equation}
This cue measures whether the candidate phase change is accompanied by a
corresponding change in the local visual representation.

The posterior and visual cues are concatenated to form a candidate-level
cue vector
\begin{equation}
\mathbf{x}_i =
\left[
\phi_i^{\mathrm{prob}}
\,\Vert\,
\phi_i^{\mathrm{vis}}
\right].
\end{equation}

The cue vector $\mathbf{x}_i$ summarizes the local evidence within the causal
candidate-centered window and is processed by a lightweight TCN followed by mean pooling, yielding the candidate representation
$\phi_i$.


A lightweight binary classifier then produces the final retain--suppress
decision
\begin{equation}
d_i
=
\mathbf{1}
\left[
\sigma
\left(
g_{\theta}(\boldsymbol{\phi}_i)
\right)
\geq
\gamma
\right],
\end{equation}
where $d_i=1$ denotes retention and event emission, whereas $d_i=0$ denotes
suppression. The threshold $\gamma$ is selected on the validation set and fixed during testing. TCV is trained with candidate labels obtained by phase-pair-aware one-to-one matching to ground-truth transitions; matched candidates are positive, while unmatched or duplicate candidates are negative.

\section{Experiments and Results}
\subsection{Experimental Setup}
\noindent\textbf{Datasets.}
We conducted experiments on the Cholec80~\cite{twinanda2016endonet} dataset, which contains 80 laparoscopic
cholecystectomy videos annotated with seven surgical phases. Following the
standard split used in our experiments, we used 32 videos for training, 8 videos
for validation, and 40 videos for testing. All videos were represented by
pre-extracted DINOv2 CLS features sampled at 1 fps.\\
\noindent\textbf{Implementation Details.}
All experiments used a frozen online phase recognizer with DINOv2 features and a causal MS-TCN. For TCE, validation selected a dwell duration of $d=5$ frames and 8 frames of pre-candidate context, which were fixed for testing. Workflow-legal phase-transition pairs were estimated exclusively from the training annotations. TCV used the resulting 13-frame causal window, combining pre/post phase-distribution cues ($\ell_1$ distance and Jensen--Shannon divergence) with pre/post DINOv2 visual change. The candidate classifier was a one-layer causal TCN with hidden dimension 32, dropout 0.1, mean pooling, and a binary output head. It was trained for 50 epochs using AdamW with batch size 128, learning rate $10^{-3}$, and weight decay $10^{-4}$. The retain threshold was selected on validation and fixed for testing. Experiments were conducted on an NVIDIA RTX 6000 Ada Generation GPU.\\
\noindent\textbf{Evaluation Protocol.}
Existing surgical phase-recognition metrics, such as frame-wise accuracy, segmental F1, and Edit Score, assess frame-level correctness and
segment-level sequence quality but do not directly measure whether discrete
workflow-transition events are emitted with the correct ordered phase pair,
without repeated triggering, and within an acceptable temporal tolerance.
We therefore evaluate emitted transitions using phase-pair-aware one-to-one
temporal matching.

Each emitted event is represented by its ordered phase pair and online
decision time,
$\hat e_i=(\hat A_i,\hat B_i,\hat t_i^{\mathrm{dec}})$.
A prediction is eligible to match a ground-truth event
$e_j=(A_j,B_j,t_j^{\mathrm{gt}})$ only if their ordered phase pairs are
identical and
$|\hat t_i^{\mathrm{dec}}-t_j^{\mathrm{gt}}|\leq\delta$,
where $\delta$ denotes the temporal tolerance. Among eligible pairs,
one-to-one matching is performed using the smallest absolute decision-time
difference.

Matched predictions are counted as true positives, unmatched predictions
as false positives, and unmatched ground-truth events as false negatives.
If multiple predictions are eligible for the same ground-truth event, only
one is matched and the remaining predictions are counted as duplicates,
which form a subset of false positives. We report event-level precision,
recall, and F1, together with
$\mathrm{False/GT}=N_{\mathrm{FP}}/N_{\mathrm{GT}}$ and
$\mathrm{Dup/GT}=N_{\mathrm{Dup}}/N_{\mathrm{GT}}$.
For matched events, online decision delay is measured as
$\hat t_i^{\mathrm{dec}}-t_j^{\mathrm{gt}}$.
All event counts are pooled across the test videos, while validation videos
are used only for threshold selection.

\subsection{Results}
Table~\ref{tab:main_results} demonstrates the progressive error reduction achieved by TTDF.
Directly emitting every hard-label change results in low precision and a
large number of false and duplicate events, confirming that raw phase
changes are not reliable transition decisions. Adding dwell persistence
substantially reduces jitter errors, decreasing False/GT
from 6.570 to 2.430 and Dup/GT from 0.332 to 0.072. Workflow legality
further removes workflow-illegal candidates, improving precision from
0.209 to 0.321 and F1 from 0.316 to 0.428 without reducing recall.

TCV subsequently suppresses remaining false candidates retained by TCE.
Relative to TCE, it reduces False/GT from 1.357 to 0.603 and Dup/GT from
0.072 to 0.013, while increasing precision from 0.321 to 0.470 and F1 from
0.428 to 0.497. Recall decreases from 0.643 to 0.531, reflecting the
expected trade-off between transition coverage and false-event suppression.
Overall, TTDF provides the best balance between event-detection performance
and transition reliability.
\begin{table}[t]
\centering
\caption{Progressive event-level transition detection on Cholec80.
TCE denotes Stage~1 with dwell persistence and workflow legality, while
TTDF additionally includes Stage~2 TCV. Delay$_{50}$ is the median online
decision delay.}
\label{tab:main_results}
\small
\setlength{\tabcolsep}{3.2pt}
\renewcommand{\arraystretch}{1.05}

\begin{tabular}{l|ccc|ccc}
\hline
\textbf{Method}
& \textbf{Prec.}$\uparrow$
& \textbf{Rec.}$\uparrow$
& \textbf{F1}$\uparrow$
& \textbf{False/GT}$\downarrow$
& \textbf{Dup/GT}$\downarrow$
& \textbf{Delay$_{50}$}$\downarrow$ \\
\hline
Raw changes
& 0.093
& \textbf{0.672}
& 0.163
& 6.570
& 0.332
& \textbf{0.0} \\
+ Dwell
& 0.209
& 0.643
& 0.316
& 2.430
& 0.072
& 5.0 \\
+ Legality (TCE)
& 0.321
& 0.643
& 0.428
& 1.357
& 0.072
& 5.0 \\
\hline
+ TCV (TTDF)
& \textbf{0.470}
& 0.531
& \textbf{0.497}
& \textbf{0.603}
& \textbf{0.013}
& 6.3 \\
\hline
\end{tabular}
\end{table}

\begin{table}[t]
\centering
\caption{Ablation of candidate-centered cues in TCV on Cholec80.
Posterior denotes the combined $\ell_1$ and JS posterior cues. Emitted Events
indicates the number of Stage-1 candidates retained by TCV. Lower is better for
Emitted Events, False/GT, and Dup/GT; higher is better for F1.}
\label{tab:cue_ablation}
\small
\setlength{\tabcolsep}{3.0pt}
\renewcommand{\arraystretch}{1.05}

\begin{tabular}{l|cccc}
\hline
\textbf{Cue}
& \textbf{Emitted Events}$\downarrow$
& \textbf{F1}$\uparrow$
& \textbf{False/GT}$\downarrow$
& \textbf{Dup/GT}$\downarrow$ \\
\hline
Posterior $\ell_1$
& 245.6$\pm$61.0
& 0.475$\pm$0.025
& 0.557$\pm$0.173
& \textbf{0.013} \\
Posterior JS
& 287.2$\pm$76.4
& 0.484$\pm$0.017
& 0.684$\pm$0.245
& 0.020$\pm$0.015 \\
Posterior $\ell_1$+JS
& \textbf{244.2$\pm$18.0}
& 0.479$\pm$0.012
& \textbf{0.551$\pm$0.047}
& \textbf{0.013} \\
Visual
& 468.0$\pm$4.5
& 0.426$\pm$0.005
& 1.354$\pm$0.008
& 0.071$\pm$0.002 \\
Posterior + visual
& 266.6$\pm$30.9
& \textbf{0.497$\pm$0.006}
& 0.603$\pm$0.094
& \textbf{0.013} \\
\hline
\end{tabular}
\end{table}
Table~\ref{tab:cue_ablation} evaluates the roles of posterior and visual cues in TCV.
Visual evidence alone retains substantially more Stage-1 candidates and
yields the highest False/GT. In contrast, posterior cues substantially reduce
emitted events and False/GT, indicating that local posterior changes provide
the primary evidence for false-candidate suppression. Among the posterior
variants, combining $\ell_1$ and JS yields the fewest emitted events and the
lowest False/GT. Adding visual evidence improves F1 from 0.479 to 0.497 while
maintaining a low duplicate rate, with a slight increase in False/GT. The combined cues therefore provide a better balance between false-candidate suppression and true-transition preservation, supporting more reliable transition detection.
\section{Conclusion}
We presented TTDF, a causal two-stage framework for reliable surgical workflow transition detection from a frozen online phase recognizer. TCE removes jitter-induced and workflow-illegal candidates through training-free filtering, while TCV suppresses persistent, workflow-consistent false candidates using phase-posterior and visual evidence. TTDF provides a  practical framework for reliable online workflow-transition event detection.

\medskip
\noindent\textbf{Acknowledgments.}
This work was supported by a PhD studentship funded by the School of Computer
Science, University of Leeds.

\medskip
\noindent\textbf{Disclosure of Interests.}
The authors have no competing interests to declare that are relevant to the
content of this article.
%
%
%
%
\bibliographystyle{splncs04}
\bibliography{references}




\end{document}